\documentclass[preprint, 10pt]{IEEEtran}

\usepackage{lineno}
\modulolinenumbers[5]
\usepackage[margin=2.5cm]{geometry}

\makeatletter
\def\ps@pprintTitle{%
 \let\@oddhead\@empty
 \let\@evenhead\@empty
 \def\@oddfoot{\centerline{\thepage}}%
 \let\@evenfoot\@oddfoot}
\makeatother

\usepackage{graphicx}
\usepackage{graphics}
\usepackage{amssymb,mathrsfs}
\usepackage{amsmath,mathtools}
\usepackage{color}
\usepackage{xspace}
\usepackage{algpseudocode}
\usepackage{bbm}
\usepackage{comment}
\usepackage{subfig}
\usepackage{psfrag}
\usepackage{lipsum}
\usepackage[utf8]{inputenc}
\usepackage[english]{babel}

\definecolor{CranJ}{cmyk}{0,0.69,0.54,0.04} 
\definecolor{PinkJ}{cmyk}{0,0.71,0.43,0.12} 
\definecolor{Cran}{cmyk}{0,0.73,0.41,0.29} 
\definecolor{VRed}{cmyk}{0,0.75,0.25,0.2} 
\definecolor{ORed}{cmyk}{0,0.75,0.75,0} 
\definecolor{CBlue}{cmyk}{1,0.25,0,0} 

 \newcommand{\reals}{{\mathbb{R}}}

\renewcommand{\baselinestretch}{.985}

\newcommand{\vect}[1]{\boldsymbol{\mathbf{#1}}}
\newcommand{\vectsf}[1]{\vect{\mathsf{#1}}}

\newcommand{\oprocendsymbol}{\hbox{$\bullet$}}
\newcommand{\oprocend}{\relax\ifmmode\else\unskip\hfill\fi\oprocendsymbol}

\title{\bf A convolution recurrent autoencoder for spatio-temporal missing data imputation}  

\author{
{\bfseries Reza. Asadi$^1$, Amelia. Regan$^2$}\\
$^1$PhD candidate, Department of Computer Science, University of California Irvine\\
$^2$Professor, Department of Computer Science, University of California Irvine\\
}

\begin{document}
\maketitle

\begin{abstract}
When sensors collect spatio-temporal data in a large geographical area, the existence of missing data cannot be escaped. Missing data negatively impacts the performance of data analysis and machine learning algorithms. In this paper, we study deep autoencoders for missing data imputation in spatio-temporal problems. We propose a convolution bidirectional-LSTM for capturing spatial and temporal patterns. Moreover, we analyze an autoencoder's latent feature representation in spatio-temporal data and illustrate its performance for missing data imputation. Traffic flow data are used for evaluation of our models. The result shows that the proposed convolution recurrent neural network outperforms state-of-the-art methods.

\end{abstract}

\vspace{1em}

\section{Introduction}
Spatio-temporal problems have been studied in broad domains \cite{atluri2018spatio}, such as transportation systems, power grid networks and weather forecasting, where data is collected in a geographical area over time. Traffic flow data are an important spatial-temporal data. Unlike traditional methods for static network flow problems \cite{asadi2016cycle} and route finding \cite{regan2016integration}, in which solving an optimization problem finds the solution, recently data-driven spatio-temporal approaches have been broadly applied on traffic flow data \cite{bae2018missing}. Spatio-temporal data are gathered by a large number of sensors and they inevitably miss observations due to a variety of reasons, such as an error prone measurements, malfunctioning sensors, or communication error \cite{li2018missing}. In the presence of missing data, the performance of machine learning tasks such as classification, clustering and forecasting drops dramatically and results in biased inference. Hence, researchers address the problem by estimating missing values in preprocessing steps, or by developing machine learning models that are robust with respect to missing data. Here we propose a method for missing data imputation in the preprocessing step.


Statistical and machine learning techniques are broadly applied for missing data imputation. The primary approach is to use an ARIMA model, which works well under linear assumptions \cite{ansley1984estimation}. A matrix completion method has also been proposed for missing data imputation \cite{yu2016temporal}; however, it requires low-rankness and static data. Dimensional reduction techniques for missing data imputation have good performance, e.g., a probabilistic principle component analysis method for missing traffic flow data \cite{qu2009ppca}, and a tensor-based model for traffic data completion \cite{ran2016tensor}. Most recently, \cite{lana2018imputation} proposes a clustering approach in spatial and temporal contexts for missing data imputation, including pattern clustering-classification and an Extreme Learning Machine with in-depth review of related work of missing data imputation in traffic flow problems. While clustering and dimensional reduction techniques differ from our model, some similarities suggests an avenue for further investigation in the future.


Increasing in the size of spatio-temporal datasets motivates researchers to develop scalable missing data imputation techniques.  Contrary to statistical techniques, neural networks do not rely on hand-crafted feature engineering and do not use prior assumptions on input data. Shallow neural networks are shown to have great performance compared with other machine learning algorithms on traffic data \cite{asadi2015rule}, but their performance reduces in large-scale problems. Recently the outperformance of deep neural networks on large-scale problems and their flexible architecture to capture spatial and temporal data illustrates their dominance over statistical and other machine learning techniques. Following the proposed denoising autoencoder with a fully connected neural network in \cite{vincent2010stacked}, a comparison of denoising autoencoders and k-means clustering for traffic flow imputation is studied in \cite{duan2016efficient}. Multiple missing data imputation with multiple training of fully connected, overcomplete autoencoders are examined in \cite{gondara2018mida}. 


Since training neural networks is computationally expensive, fully-connected, multiply trained and overcomplete autoencoders can be inefficient solutions for large scale problems. Moreover, recent works demonstrate the increased performance of convolutional layers and LSTM layers for extracting spatial and temporal patterns compared to fully connected layers. A Convolutional neural network is proposed for missing data imputation in traffic flow data \cite{zhuang2018innovative}. The model captures spatial and short term patterns with a convolutional layer. A bidirectional LSTM with a modification on the LSTM neurons is proposed \cite{cao2018brits}, but spatial data is not considered. Convolutional recurrent neural networks have great performance in large-scale spatio-temporal problems \cite{asadi2019spatial}. In \cite{jia2017spatio}, a spatio-temporal autoencoder is proposed for high dimension patient data with missing values, and classifiers are used for at top of feature learning.

In the aforementioned works deep neural networks have been studied on spatio-temporal data. However, there is lack of analysis in applying convolution-recurrent autoencoders on spatio-temporal problems for missing data imputation in traffic flow problems with the objective of learning spatial patterns with convolution and temporal patterns with LSTM layers. In this paper, we first propose a convolution recurrent autoencoder for multiple missing data imputation. The model is examined on traffic flow data. It is shown that the proposed convolution recurrent autoencoder improves the performance of missing data imputation problem. Moreover, the latent feature representation of the autoencoders is analyzed. This analysis shows that the latent feature space is semantically meaningful representation of traffic flow data. We also examine the performance of applying $\mathsf{k}$-nearest-neighbor (KNN) to evaluate the effectiveness of using autoencoders' latent representation in missing data imputation. The proposed model can be applied for missing data imputation in spatio-temporal problems.

\section{Preliminaries}
\subsection{Problem definition}
\begin{figure}[t]
\includegraphics[width=6cm]{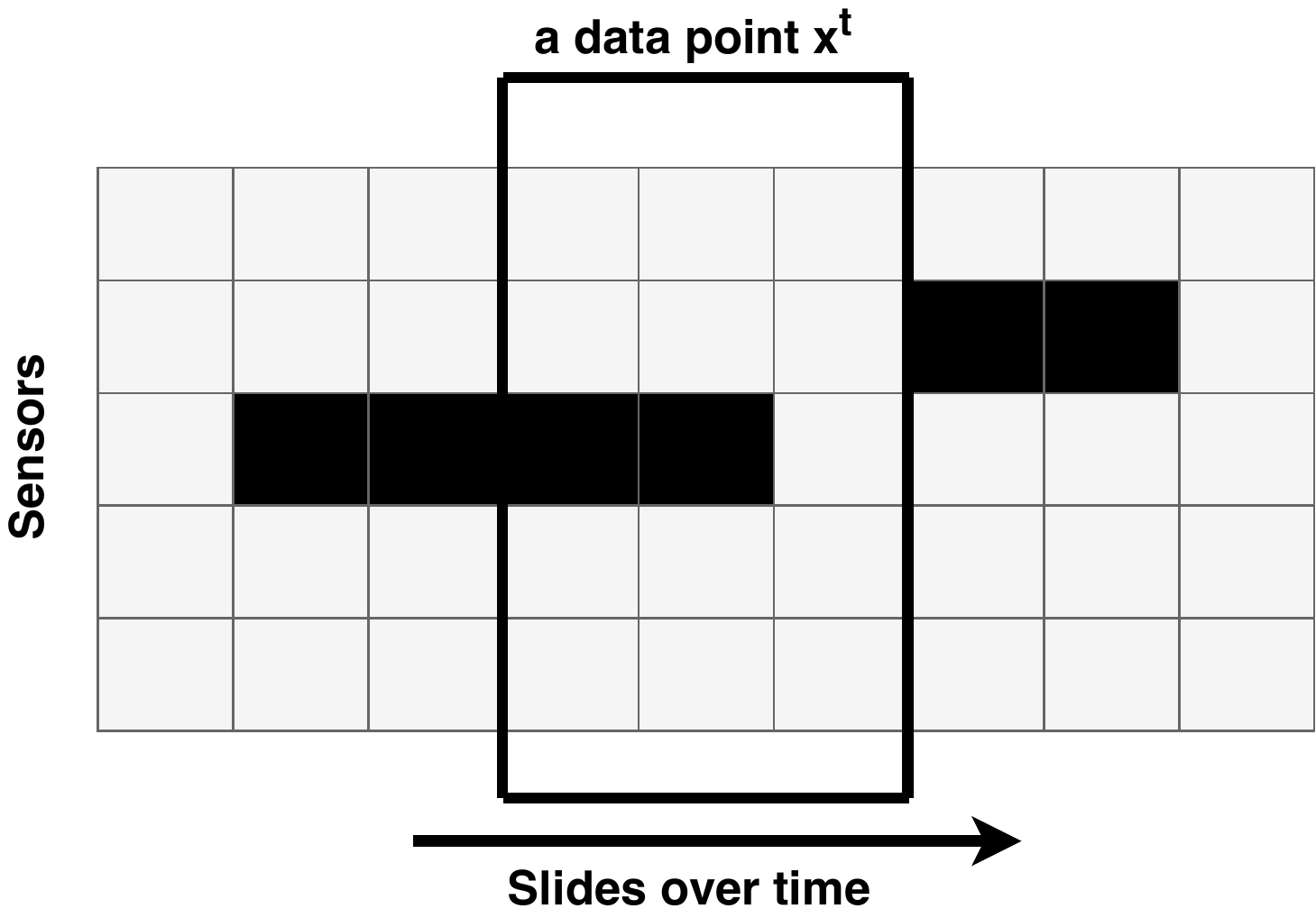}
\centering
\caption{A sliding window selects subsamples and feeds these into an autoencoder. The missing values are represented in black.}\label{fig::probDef}
\end{figure}

Spatio-temporal data is represented by a matrix $\vectsf{X} \in \reals^{\mathsf{s} \times \mathsf{\bar{t}} \times \mathsf{f}}$, where $\mathsf{s}$ is the number of sensors, $\mathsf{\bar{t}}$ is the number of time steps and $\mathsf{f}$ is the number of features. Missing data can exist in various ways, for example at individual points or over intervals, where one sensor loses data for a period of time. To apply a deep neural network for time series imputation, a sliding window method generates $\vectsf{x}^\mathsf{t} \in \reals^{\mathsf{s} \times \mathsf{w} \times \mathsf{f}}$, where $\mathsf{w}$ is time window and $\mathsf{t} \in [0, \mathsf{\bar{t}}]$. In the rest of the paper, we call $\vectsf{x}^\mathsf{t}$ as a data point. For the purpose of training and evaluation, an interval of missing values is added to the input data and represented with $\vectsf{x}_\mathsf{m}^\mathsf{t}$. The objective is to impute missing values for $\vectsf{x}_\mathsf{m}^\mathsf{t}$ using spatial and temporal correlation. In Fig. \ref{fig::probDef}, a schematic example of applying a sliding window on a spatial time series with interval-wise missing values is represented.

\subsection{A denoising autoencoder}
An autoencoder decoder $\mathsf{AD}(.)$ proposed in \cite{vincent2010stacked} and can be applied in missing data imputation problem. In the training process, a denoising encoder decoder receives $\vectsf{x}_{\mathsf{m}}^\mathsf{t}$ as input and $\vectsf{x}^\mathsf{t}$ as target data. It reconstructs its input $\vectsf{\bar{x}^\mathsf{t}} = \mathsf{AD}(\vectsf{x}^\mathsf{t}_\mathsf{m})$ by minimizing the loss function $\mathsf{Loss}(\vectsf{x}^\mathsf{t}, \vectsf{\bar{x}}^\mathsf{t})$, e.g. mean square loss function, \cite{vincent2010stacked}, for autoencoders' output $\vectsf{\bar{x}^\mathsf{t}}$. In other words, the autoencoder receives a data point with some missing values and reconstructs it with the objective of accurate missing data imputation. An encoder reduces the dimension to a latent feature space $\vectsf{h} \in \reals^\mathsf{d}$, where $\mathsf{d} < \mathsf{n}$, which extracts the most important patterns of the input data. An autoencoder is capable of producing semantically meaningful representations on real-world datasets \cite{le2011building}. The decoder reconstructs the input from its latent representation. For a two layer encoder decoder, an encoder is represented with $\vectsf{h} = \sigma(\mathsf{drop}(\vectsf{x})\vectsf{w}^1+\vectsf{b}^1)$ and a decoder is represented with $\vectsf{\bar{x}} = \sigma(\mathsf{drop}(\vectsf{h})\vectsf{w}^2+\vectsf{b}^2)$, where $\sigma(.)$ is the activation function and $\mathsf{drop}(.)$ is dropout function. A multi layer fully connected, convolution or recurrent layers can be used as an encoder or decoder.



\section{A convolution-recurrent deep neural network encoder decoder framework}

\begin{figure}[t]
\includegraphics[width=8cm]{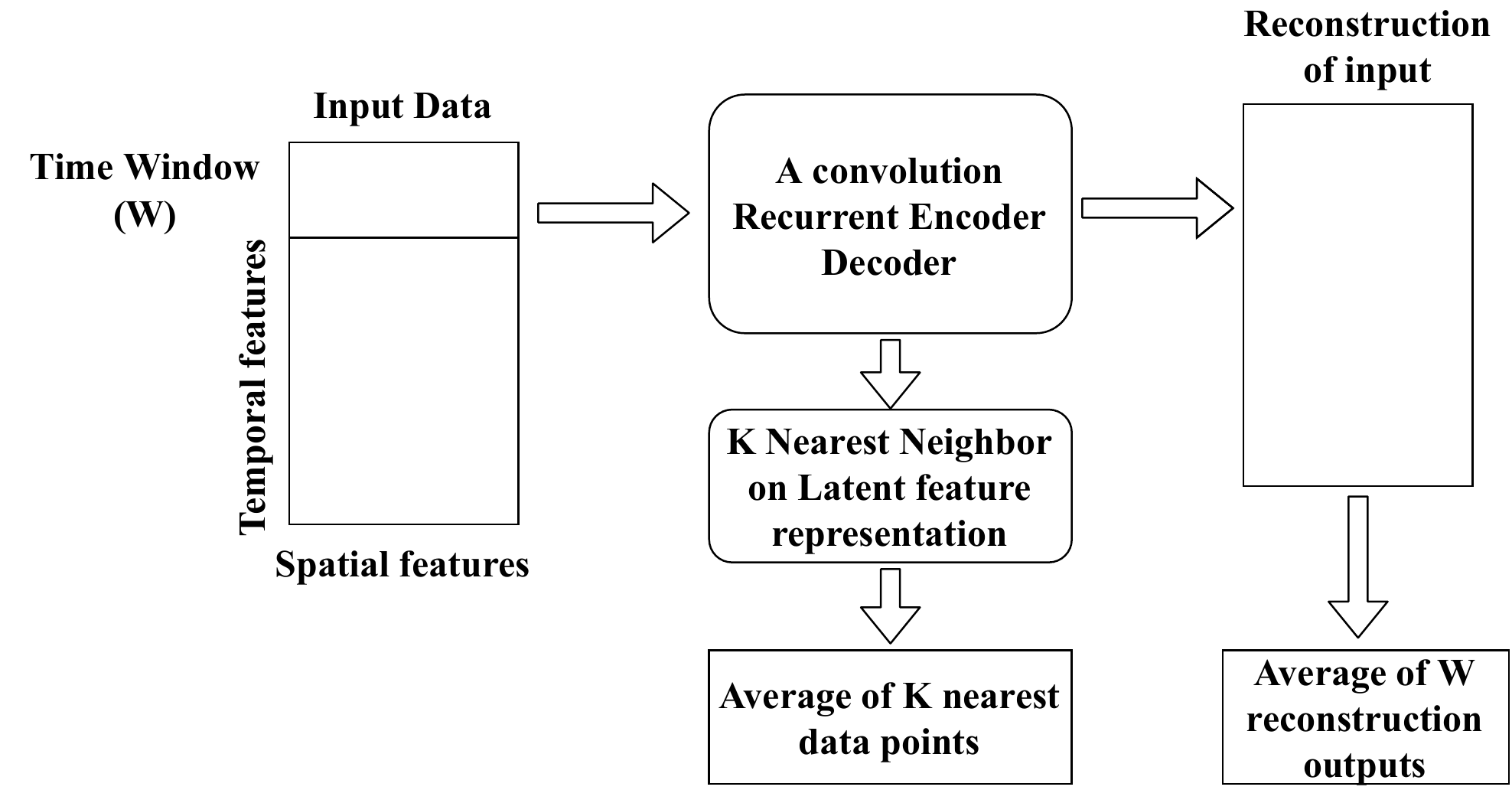}
\centering
\caption{The framework for multiple imputation with autoencoders}\label{fig::framework}
\end{figure}

As discussed in detail in \cite{schafer1998multiple}, in multiple imputation each missing datum is replaced with weighted average of more than one imputations. Hence, we propose a framework for multiple missing data imputation on spatio-temporal data, represented in Fig. \ref{fig::framework}. 

A sliding window method gives the input data with size of $\mathsf{w}$ to the autoencoder. A convolution recurrent autoencoder reconstructs the input data and automatically imputes missing values. There are $\mathsf{w}$ reconstructed values for each time window. The average of these reconstructed values is the output of neural network. The evaluation of reconstructed values is shown in Section. \ref{sec::compresults}. The second approach is with the latent feature representation of autoencoders. A KNN finds the most similar $\mathsf{k}$ data points in training data. The average of these produces the imputed values for the testing data. The model is evaluated in Section. \ref{sec::latent}.

\subsection{A CNN-BiLSTM Autoencoder}\label{sec::method1}
\begin{figure}[t]
\includegraphics[width=8cm]{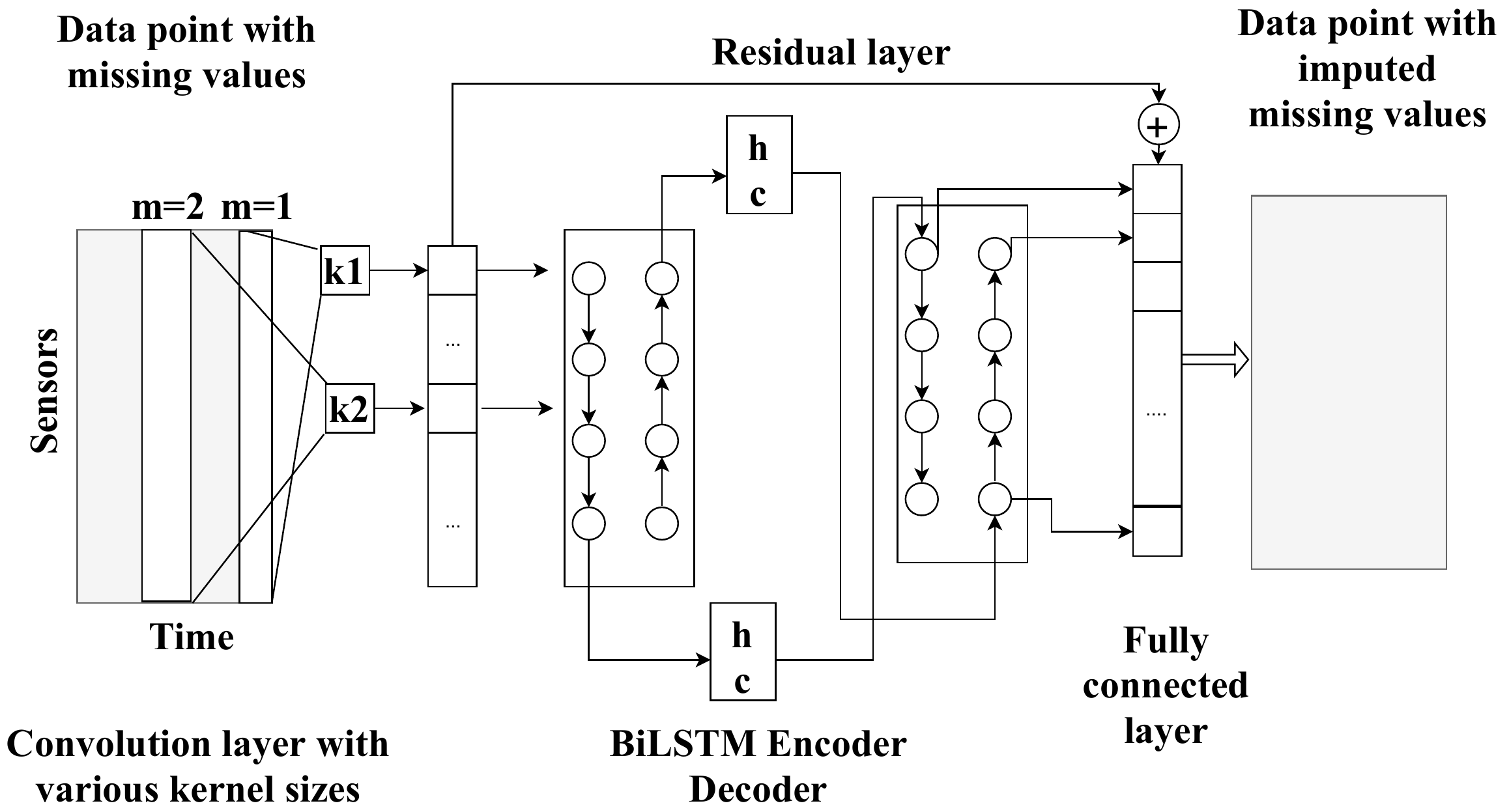}
\centering
\caption{A convolutional BiLSTM encoder decoder (CNN-BiLSTM-Res) for missing data imputation}\label{fig::model}
\end{figure}

Here we introduce the proposed convolution recurrent autoencoder for spatio-temporal missing data imputation. The proposed model is illustrated in Fig. \ref{fig::model}.

To extract spatial and temporal patterns, an encoder consists of both convolution and LSTM layers. A convolutional layer has a kernel, which slides over spatial time series data $\vectsf{X} \in \reals^{\mathsf{s} \times \mathsf{w} \times \mathsf{c}}$, where $\mathsf{c}$ is the number of channels. For non-grid data, sliding a kernel on spatial features loses the network structure and reduces the performance of model \cite{asadi2019spatial}. Hence the kernel only slides over the time axis. The kernel size is $(\mathsf{s}, \mathsf{m})$, where $\mathsf{m}<\mathsf{w}$, and stride size is $(\mathsf{s}, 1)$. Various length of kernel have been shown to have better performance. Hence, several kernels with different values of $\mathsf{m}$ are applied to the input data. The output of each kernel $\mathsf{i}$ is $\vectsf{k}^\mathsf{i} \in \reals^{1 \times \mathsf{w} \times \mathsf{f}}$, where $\mathsf{f}$ is the filter size. All of the outputs are concatenated and represented with $\vectsf{h} \in \reals^{1 \times \mathsf{w} \times \mathsf{F}}$, where $\mathsf{F}$ is the total size of all filters.

An LSTM layer receives the output of convolution layer, represented by $\vectsf{X} \in \reals^{\mathsf{w} \times{F}}$. An LSTM cell uses input, output and forget gates to prevent vanishing gradients in recurrent cells. It also returns hidden state $\vectsf{h}^\mathsf{t}$ and cell state $\vectsf{c}^\mathsf{t}$. A bidirectional LSTM layer captures the relation of past and future data simultaneously. It has two sets of LSTM cells which propagate states in two opposite directions. Thus a bidirectional LSTM layer is used for the recurrent component. Given $\mathsf{l}_1$ as the number of units in LSTM layer, the output of bidirectional LSTM is $\vectsf{h} \in \reals^{\mathsf{w} \times 2 \times \mathsf{l}_1}$. The latent feature representation of encoder consists of LSTM states $[\vectsf{h}_\mathsf{forward}^\mathsf{t}, \vectsf{c}_\mathsf{forward}^\mathsf{t},\vectsf{h}_\mathsf{back}^\mathsf{t}, \vectsf{c}_\mathsf{back}^\mathsf{t}]$, where these are the hidden and cell states of the forward and backward direction of bidirectional LSTM.

The decoder receives the encoder states and encoder output. The decoder consists of a bidirectional LSTM and a fully connected layer. The LSTM layer receives the hidden and cell states of the encoder to reconstruct the input data. A bidirectional model reconstructs past and future data. It follows with a fully connected layer with linear activation function.

Training the encoder decoder with convolution and LSTM layers is slow, as the gradient of the loss function is propagated backward on to LSTM cells and then convolutional layers. To increase the speed of training, we used a residual layer, introduced in \cite{he2016deep}, to connect the output of the convolution layer to the fully connected layer with a $\mathsf{Add}(.)$ function. In the training process, the convolution layer receives more effect from the gradient of loss function and as a result, there is faster convergence for the encoder decoder to learn spatial and temporal patterns.

The reconstruction of input automatically imputes missing data from the spatial and temporal correlation among neighboring areas. Given a time window $\mathsf{w}$, every time stamp $\vectsf{x}^\mathsf{t}$ is reconstructed $\mathsf{w}$ times and the average is used for missing imputation. An autoencoder decoder reconstructs input data  $\vectsf{\bar{x}}^\mathsf{t} = \mathsf{AD}(\vectsf{x}_\mathsf{m}^\mathsf{t})$ by minimizing loss function $\mathsf{Loss}(\vectsf{\bar{x}}^\mathsf{t}, \vectsf{x}^\mathsf{t})$ for all time steps $\mathsf{t}$. 

\subsection{Missing data imputation using latent feature representations}
A KNN algorithm compares the distance among all data points, and finds the $\mathsf{k}$ nearest data points. This approach find the most similar data points and then find the average for missing data imputation. With a sliding window approach, the number of data points in training data is the same as number of time steps $\mathsf{\bar{t}}$. For a given data point $\mathsf{x}^\mathsf{t}$ which is a matrix of size $\mathsf{len} = \mathsf{s} \times \mathsf{w} \times \mathsf{f}$, the total number of comparison in KNN is $\mathsf{t}^2 \times \mathsf{len}$. Moreover, a time series distance can be obtained with Dynamic Time Warping \cite{salvador2007toward}, which is computationally more expensive than euclidean distance.

The latent representation of autoencoder is a fixed size and reduced dimension vector $\vectsf{h} \in \reals^\mathsf{d}$. Applying KNN on latent representation is computationally more efficient than on time series data points. The total comparison is  $\mathsf{t}^2 \times \mathsf{d}$ and the latent feature distance can be computed with euclidean distance, faster than Dynamic Time Warping. In the experimental analysis, we evaluate the computational time of applying KNN on latent feature. Moreover, the average of $\mathsf{k}$ most similar data points is used as multiple missing imputation. The results of this analysis is compared with PCA-KNN in experimental results.

\section{Experimental results}

\subsection{Dataset}
\begin{figure}[t]
\includegraphics[width=6cm]{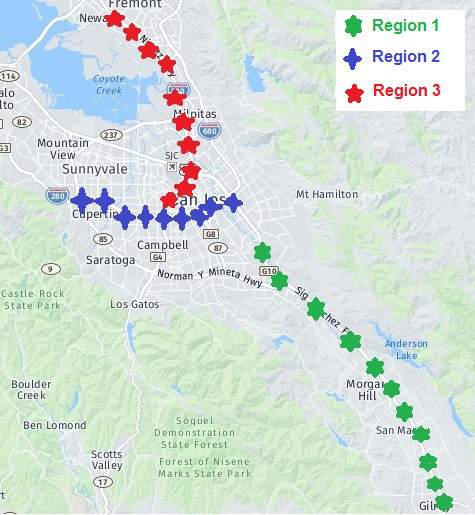}
\centering
\caption{Three regions of highways are selected for missing data imputation analysis.}\label{fig::map}
\end{figure}

We examine the performance of the proposed model on traffic flow data available in PeMS \cite{californiapems}. Traffic data are gathered every 30 seconds and aggregated every 5 minutes using loop detector stations on highways. We use three subset of stations in the Bay Area, represented in Fig. \ref{fig::map}, and evaluate the average performance of our model on these three regions to have better evaluation of the models. The first region has 10 mainline stations on 22 miles of highway US 101-South. The second region has 9 mainline stations on 10 miles of I-280-South highway, and the third region has 11 mainline stations on 13 miles of I-880-South. The training data is for the first 4 months of 2016 and the testing data is for next 2 months. The selected sensors have more than 99\% of available data for this time period.

\subsection{Preprocessing}
The data is scaled to range of [0-1] where for each data set, 0 is the minimum flow observed and 1 is the maximum. A sliding window approach is used to generate image-like input for time series data. During the experiments, we found out a time window of size 6, 30 minutes, works well. Each data point is represented with $\vectsf{X}^\mathsf{t} \in \reals^{\mathsf{s} \times 6 \times 1}$, where $\mathsf{s}$ is the number of sensors for each region.

To evaluate the model for missing data imputation, we added missing blocks to have a ground truth for evaluation. The missing data is generated randomly on training and testing data. We generated blocks of missing data with size of 0.5 to 4 hours. The sensors are randomly selected for each missing block. In the analysis, training data without missing values cannot result in a robust autoencoder for missing data imputation. Therefor, 25\% percent of training and testing data is considered as missing values. In the analysis, the performance of missing data imputation models are examined only on these missing blocks, represented with index list of $\vectsf{I}_{\mathsf{m}}^{\mathsf{test}}$.

\subsection{Baseline missing data imputation models}
Our first missing data imputation method uses a temporal average to fill missing data. Traffic flow patterns are repeated every week. Hence, a weekly-hourly average table is obtained from training data (W-H-Average). The main drawback of using temporal average is that specific days such as holidays or event days (games, festivals, concerts) have their own patterns and they are not repeated in the training data.

The second method uses the closest sensors to estimate the missing data. The value of traffic flow should be similar to the closest sensors on highways. Following the work \cite{asadi2019spatial}, a Dynamic Time Warping distance method finds the most similar sensors using time series residuals. The method uses the average of the two closest sensors and estimates the missing data (Neighbor-Value).

In the third baseline method, the most important principle components are selected, then a KNN finds the most similar data points. The average of $\mathsf{k}$ nearest values is used to estimate missing data (KNN-PCA). In the analysis, we examine different values of PCA components. The first 10 components contain more than 95 \% information ratio. Also, larger values of $\mathsf{k}$, improves the result, as the average of several missing imputations is usually a better estimation for missing values. The best size of PCA components and $\mathsf{k}$ are 10 and 20, respectively. The number of features is the number of sensors multiplied by time window, which is 60, 54 and 66 for three regions. The best values of MAE and RMSE are shown in Table. \ref{tab::results}.

\subsection{Autoencoder models}

Here we describe the implemented autoencoders. For all of the models, the batch size is set to 256 and the epochs are set to 100. An ADAM optimizer with learning rate of 0.001 is used for training the model.

A fully connected denoising encoder decoder is implemented for missing imputation FC-NN. The model is trained with architecture of (32, 16, 12, 16, 32) obtained by grid search over various number of layers and hidden units. Each layer is a fully connected layer with a Leaky-RELU activation function.

\begin{table}[t]
\begin{tabular}{ |p{2.5cm}||p{2.0cm}|p{2.0cm}|  }
 \hline
 \multicolumn{3}{|c|}{Missing data imputation error for traffic flow data} \\
 \hline
Models & MAE  & RMSE\\
 \hline
  \hline
 W-H-Average   &   26.3  & 34.8 \\
  \hline
Neighbor-value   &  38.9   &  45.5 \\
 \hline
KNN-PCA   &  19.0   &  25.5  \\
 \hline
FC-NN   &  14.3   & 21.5  \\
 \hline
LSTM  &   10.1  & 16.0 \\
 \hline
BiLSTM   &  7.8   & 14.0 \\
 \hline
CNN-BiLSTM  & 7.6   & 13.9\\
 \hline
CNN-BiLSTM-Res  & \textbf{6.8}   & \textbf{13.0}\\
 \hline
 \end{tabular}
 \caption{The comparisons of missing data imputation results}\label{tab::results}
\end{table} 

To capture temporal patterns, an LSTM encoder decoder with 32 neurons is trained LSTM. To capture the effect of past and future data points, a bidirectional LSTM is implemented with 16 neurons in each direction BiLSTM. A dropout with parameter 0.2 prevents over-fitting the LSTM layers. A convolution recurrent encoder decoder CNN-BiLSTM is implemented with four kernels of size $(\mathsf{s},1)$, $(\mathsf{s},2)$, $(\mathsf{s},3)$ and $(\mathsf{s},4)$ and filter size of 8 and a Leaky Relu activation function. The bidirectional-LSTM has 16 units on each direction and is connected to a fully connected layer with the size of input sensors. Slow convergence of convolutional-BiLSTM model motivates us to add residual layer connecting convolution to the output of BiLSTM for faster gradient propagation. The model CNN-BiLSTM-Res, the proposed model in Fig. \ref{fig::model}, is with the same architecture of CNN-BiLSTM but with residual layer. All implementations has been done with Tensorflow and Keras \cite{abadi2016tensorflow}.

\subsection{Comparison of results}\label{sec::compresults}
Given $\vectsf{x}$ as real value and $\bar{\vectsf{x}}$ as predicted value, Mean Absolute Error (MAE), and Root Mean Square Error (RMSE) are used for evaluation. Given a set of missing data points in testing data $\vectsf{X}_{\mathsf{m}}^{\mathsf{test}}$ and their corresponding indices $\vectsf{I}_{\mathsf{m}}^{\mathsf{test}}$, the index $\mathsf{i}$ is selected from the index set of missing data $\vectsf{I}_{\mathsf{m}}^{\mathsf{test}}$ in \ref{eq::eval1} and \ref{eq::eval2}.

\begin{align}\label{eq::eval1}
    \mathsf{MAE} &= \frac{1}{\mathsf{n}} \sum_{\mathsf{i} \in \vectsf{I}_{\mathsf{m}}^{\mathsf{test}}} |\vectsf{x}_\mathsf{i} - \bar{\vectsf{x}}_\mathsf{i}|\\
    \mathsf{RMSE} &= \sqrt{\frac{1}{\mathsf{n}} \sum_{\mathsf{i} \in \vectsf{I}_{\mathsf{m}}^{\mathsf{test}}} (\vectsf{x}_\mathsf{i} - \bar{\vectsf{x}}_\mathsf{i})^2} \label{eq::eval2}
\end{align}

\begin{figure}[t]
\includegraphics[width=8cm]{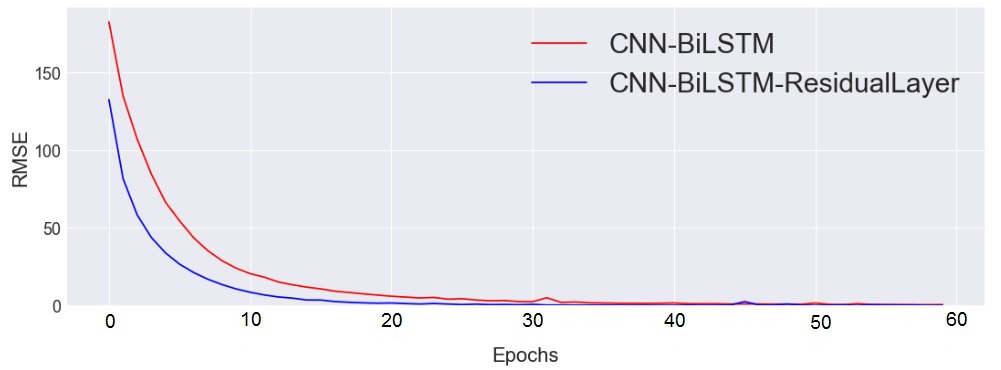}
\centering
\caption{The comparison of validation loss during training of autoencoder models}\label{fig::residuallayer}
\end{figure}

The results are represented in Table. \ref{tab::results}. It shows that the temporal and spatial averages, the first two models have a poor performance for missing data imputation. Among three baseline models, KNN-PCA is the best missing data imputation technique. Autoencoders have significantly better performance than baseline models. The LSTM model has good performance for missing data imputation compared with FC-NN for capturing temporal patterns. A bidirectional LSTM shows great performance by capturing the relation between past and future data simultaneously. A CNN-BiLSTM hardly converges to the optimum solution but is not better than the BiLSTM model. Finally, the proposed CNN-BiLSTM-Res encoder decoder has the best MAE and RMSE. It shows that a residual layer improves the performance for a combination of convolution and LSTM layers. The model CNN-BiLSTM-Res has 13\% and 7\% improvement on MAE and RMSE compared with the best BiLSTM model. As it is illustrated in Section. \ref{sec::method1}, because of the slow convergence of convolutional LSTM models, a residual layer is used to propagate gradients of loss function directly to convolution layer. In Fig. \ref{fig::residuallayer}, the convergence of CNN-BiLSTM and CNN-BiLSTM-Res are represented, which shows faster convergence of CNN-BiLSTM-Res.

\begin{figure}[t]
\includegraphics[width=8cm]{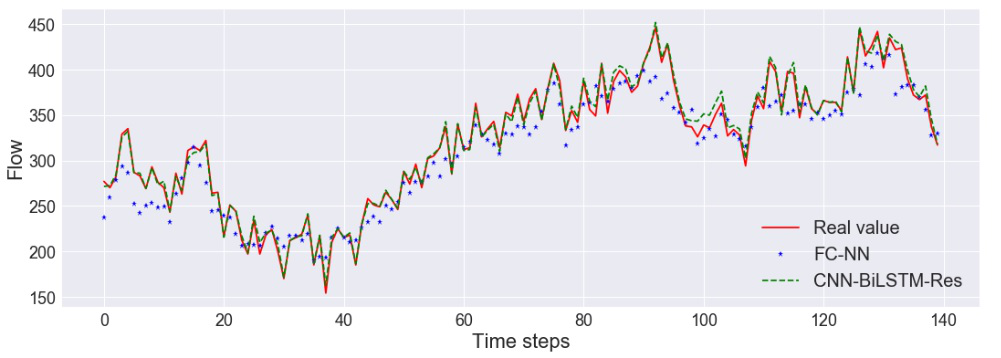}
\centering
\caption{The comparison of missing data imputation models for one interval of missing values}\label{fig::plotResults}
\end{figure}

\begin{figure}[t]
\includegraphics[width=8cm]{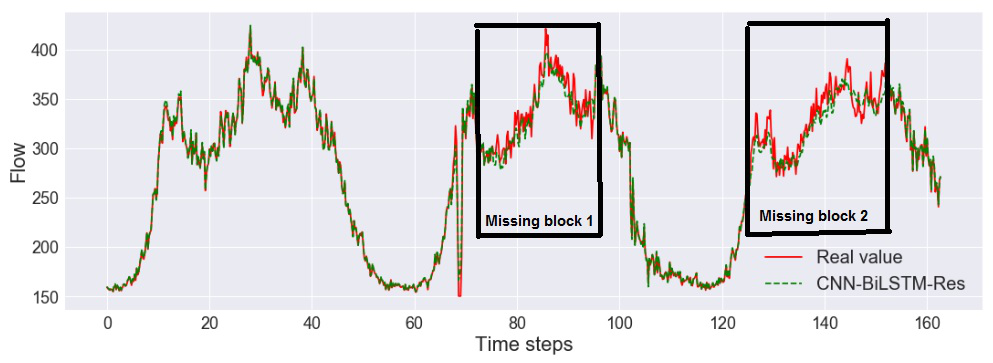}
\centering
\caption{The illustration of missing data imputation for one sensor by the proposed model}\label{fig::plotResults-week}
\end{figure}

In Fig. \ref{fig::plotResults}, the prediction results is represented for FC-NN and CNN-BiLSTM-Res as the example of missing data imputation results. Compared with FC-NN, the prediction result of CNN-BiLSTM-Res is clearly more accurate missing imputation and closer to ground truth. In Fig. \ref{fig::plotResults-week}, the plot illustrates the missing data imputation by CNN-BiLSTM-Res for two missing blocks during three days, and shows the closeness of imputed data to real traffic flow data. This output example shows the estimation of missing block of data is very close to real values; however, still the distance between real and predicted values for missing blocks is more than healthy data, which are the time series values out of missing blocks.

\subsection{Discussion on multiple missing data imputation}

For non-temporal data, an autoencoder reconstructs one value for each input data point. However, for temporal data, a sliding window generates data points for each time step. Referring to Figure \ref{fig::probDef}, the data point actually contains all of the values within a time window. For a given time window $\mathsf{w}$, there are $\mathsf{w}$ reconstructed values for each time step. The result in Table. \ref{tab::results} is for $\mathsf{w}$ multiple missing data imputation. Here we use one step reconstruction of each output for comparison purpose. In other words, here we describe a single missing imputation output of applying autoencoders on traffic flow data.

The value of MAE for FC-NN, LSTM, BiLSTM and CNN-BiLSTM are 23.7, 15.5, 11.9, 6.9, respectively. Also, the RMSE for FC-NN, LSTM, BiLSTM and CNN-BiLSTM are 32.1, 22.5, 18.1, 13.7, respectively. Comparing to Table. \ref{tab::results}, we can see that a single missing data imputation has very lower performance. The analysis shows that multiple imputation and using the average of them significantly improves missing data imputation. This multiple imputation approach improves the output of autoencoders on time series data.

\subsection{Latent feature representation}\label{sec::latent}

\begin{figure}[t]
\includegraphics[width=5.1cm]{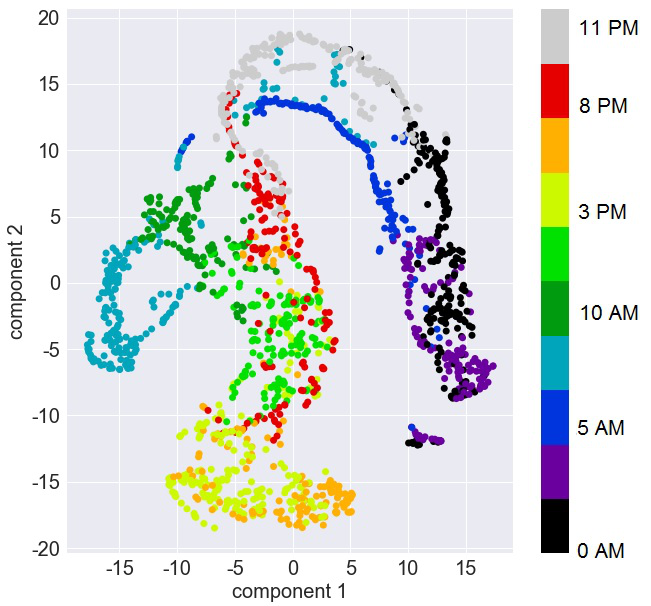}
\centering
\caption{The latent feature space visualization of FC-NN with t-SNE. Each data point has a color that represents the time of day.}\label{fig::latent}
\end{figure}

\begin{figure}[t]
\includegraphics[width=8cm]{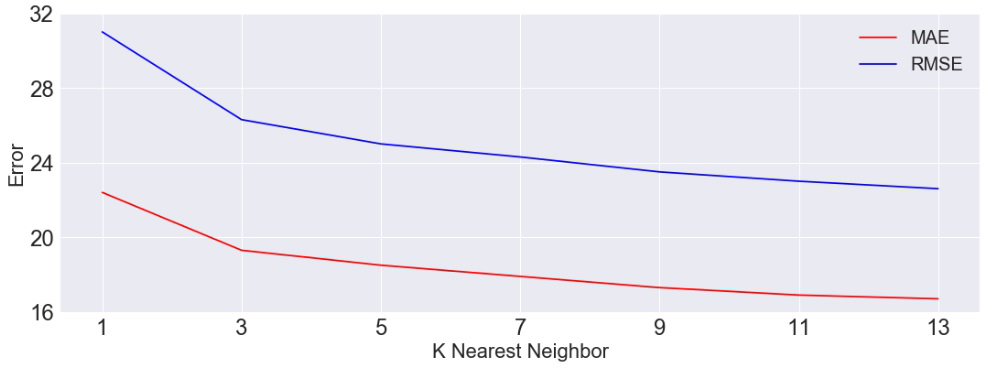}
\centering
\caption{The comparison of applying KNN on FC-NN latent feature for various size of $\mathsf{k}$}\label{fig::ksize}
\end{figure}

\begin{figure}[t]
\includegraphics[width=8cm]{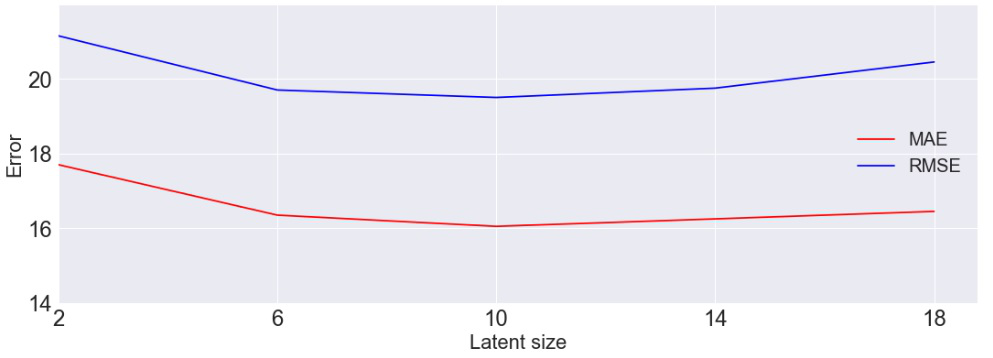}
\centering
\caption{The comparison of applying KNN on FC-NN latent feature for various size of latent features}\label{fig::latentsize}
\end{figure}

The latent feature representation of autoencoders illustrates meaningful information. In Fig \ref{fig::latent}, a t-SNE method \cite{maaten2008visualizing} visualizes latent feature representation of hidden state of (FC-NN) for 7 days. The plot shows that for each time of day the patterns of data points are closer to each other. Here our objective is to illustrate how latent feature representation can be used for missing data imputation. Hence we use the concept of similarity of data points. A KNN is applied on latent feature representation in training data points. The $\mathsf{k}$ most similar data points are used. The error for the average of different values of $\mathsf{k}$ is represented in Fig \ref{fig::ksize}. The plot shows that a 1 nearest neighbor on latent feature representation results in 23.5 and 31.0 for MAE and RMSE scores. However, a 13 nearest neighbor results in 16.7 and 22.6, MAE and RMSE, respectively. The reduction in missing data imputation error shows the effectiveness of multiple imputation on latent feature representation. We also examine the relation between size of latent features and missing imputation on FC-NN in Fig. \ref{fig::latentsize}. The analysis shows that across latent sizes of 2 to 20 there are changes in the performance of the missing data imputation. The results suggest that the best latent size is 10.

A KNN is applied on latent feature representation of various implemented autoencoders. The results of applying KNN on latent feature of FC-NN, hidden and cell state of LSTM and BiLSTM have MAE of 16.6, 18.1, 17.8 and RMSE of 22.5, 24.1, 23.8, respectively. While a FC-NN with six layers is the best model to generate latent features, the other convolution-recurrent models cannot easily generates a latent feature representation for missing data imputation. One conclusion is that size of latent vector greatly effect on the result. A KNN on smaller size of latent vector finds better missing data imputation. The analysis also shows that applying KNN on the latent feature of FC-NN is better than KNN-PCA, which shows autoencoders are capable of generating better latent feature representation for traffic flow data.


\section{Conclusion and Future Work}

In this paper, we study autoencoders for missing data imputation in spatio-temporal problems and examined the performance of various autoencoders to capture spatial and temporal patterns. We illustrate that a convolution recurrent autoencoder can capture spatial and temporal patterns and outperforms state-of-the-art missing data imputation. We conclude that a convolution layer with various kernel sizes and a bidirectional LSTM improves missing data imputation in traffic flow data. Also, the slow convergence of the convolution-recurrent autoencoder is improved with a residual layer. We also describe an approach considering multiple imputation for autoencoders for time series data. The results show that multiple imputation is significantly better than single imputation. Moreover, We illustrate advantage of using the latent feature of autoencoders for missing data imputation. We describe an approach for using autoencoders' latent feature representation for multiple imputation. The analysis shows that it outperforms KNN on principle components of traffic flow data. However, the latent feature of convolution-recurrent autoencoders needs a careful design of the architecture to obtain better results and can be explored more in future works.

Future research will focus on generative neural networks. Moreover, while it is shown that convolution-recurrent neural networks show a great performance for spatio-temporal problems, spatial and temporal clustering techniques can make the model more effective on larger geographical areas.

\bibliography{mybibfile}

\end{document}